%% file: integrating_information_cima.tex
\documentclass{ecai}
\usepackage{times}
\usepackage{graphicx}
\usepackage{latexsym}

\input{master_include/master_preamble}

\usepackage{tabularx}
\usepackage{subcaption}
\usepackage{nameref}

\newcommand{\citet}[1]{\cite{#1}}
\newcommand{\citep}[1]{\cite{#1}}

\begin{document}

\title{Experimental and causal view on information integration in autonomous agents}

\author{Philipp Geiger\institute{Max Planck Institute for Intelligent Systems, T\"ubingen, Germany; email: philipp.geiger@tuebingen.mpg.de} \and Katja Hofmann\institute{Microsoft Research Cambridge, Cambridge, United Kingdom; email: katja.hofmann@microsoft.com} \and Bernhard Sch\"olkopf \institute{Max Planck Institute for Intelligent Systems, T\"ubingen, Germany; email: bs@tuebingen.mpg.de} }

\maketitle
\bibliographystyle{ecai}

\begin{abstract}
The amount of digitally available but heterogeneous information about the world is remarkable, and new technologies such as self-driving cars, smart homes, or the internet of things may further increase it.
In this paper we examine certain aspects of the problem of how such heterogeneous information can be harnessed by autonomous agents.
After discussing potentials and limitations of some existing approaches, we investigate how \emph{experiments} can help to obtain a better understanding of the problem.
Specifically, we present a simple agent that integrates video data from a different agent, and implement and evaluate a version of it on the novel experimentation platform \emph{Malmo}.
The focus of a second investigation is on how information about the hardware of different agents, the agents' sensory data, and \emph{causal} information can be utilized for knowledge transfer between agents and subsequently more data-efficient decision making.
Finally, we discuss potential future steps w.r.t.\ theory and experimentation, and formulate open questions.
\end{abstract}

\section{Introduction}

Increasing amounts of heterogeneous information are recorded and connected, and this trend is likely to continue in the light of new technology such as self-driving cars, smart homes with domestic robots, or the internet of things.
Intuitively, it makes sense to design autonomous agents in a way that they automatically integrate all relevant and well-structured information on their environment that is available.
Various aspects of the problem of designing such agents have been investigated previously.
In this paper we approach the problem from two directions which, to our knowledge, have not been (exhaustively) examined yet: 
using sophisticated simulated experiments, on a practical level, and causal models, on a more theoretical level.
The complexity of the problem allows us only to take small steps.

 


%


\subsection{Main contributions}


The main contribution of this paper consists of two investigations: 
\begin{itemize}
\item In Section \ref{sec:aix_exp} we use a simulated \emph{experimentation platform Malmo} to obtain a \emph{better understanding} of the problem of integrating heterogeneous information. 
More specifically, we present a simple agent that harnesses video data from a different agent, and implement and evaluate a version of it. 

\item In Section \ref{sec:causal} we investigate how detailed information on the \emph{hardware of different agents} (we consider self-driving cars as example), their sensory data, and physical or \emph{causal information} can be utilized for knowledge transfer between them and subsequent more data-efficient decision making.
\end{itemize}
The common structure of both investigations is that we start with a description of a scenario that captures certain core aspects of the general problem, in particular containing a variety of heterogeneous information sources, and then sketch a method to perform information integration and subsequent decision making in these scenarios.
After experimentally evaluating the method, or illustrating it based on a toy example, we conclude both investigations with a discussion of the advantages and limitations of the respective methods.

A reoccurring theme in our investigations is that we try to treat as much information (including models) as possible explicitly as input to algorithms instead of implicitly encoding it into algorithms.
Our hope is that this sheds a better, more explicit light on the problem.

\subsection{Structure of the paper}

The paper is structured as follows:
We introduce the experimentation platform and basic concepts in Section \ref{sec:pre}.
In Section \ref{sec:problem}, we formulate the problem.
In Section \ref{sec:related}, investigate potentials and limitations of existing approaches for the problem.
In Sections \ref{sec:aix_exp} and \ref{sec:causal}, we present our two main investigations.
In Section \ref{sec:future}, we discuss future directions and pose open questions.
We conclude with Section \ref{sec:conclusions}.

\section{Preliminaries}
\label{sec:pre}

Here we introduce the concepts, models and the experimentation platform we will use in the paper.

\paragraph{Autonomous agents.}

By an \defi{(autonomous) agent} we mean a mechanism which, at each time point $t$, takes some input from the \defi{environment}, in particular its sensory data which we refer to as \defi{observation}, and outputs some \defi{action} that influences the environment. 
Moreover, by an \defi{intelligent (autonomous) agent} we mean an autonomous agent which is successful in using its inputs and outputs for given \defi{tasks}, i.e., specific goals w.r.t.\ the environment, often encoded by a \defi{reward} or \defi{utility} function.

Note that in this paper we do not define a clear boundary between agent and environment. 
Usually, we consider the hardware platform of an agent (e.g., the car) as part of the agent. 
This particularly has to be kept in mind when we talk of several agents in the ``same environment'': the hardware of the agents may still differ.
(It is almost a philosophical problem to define what precisely the ``same environment'' means. Here we simply suggest to interpret this notion as if it were used in an everyday conversation.
We pose a related question in Section \ref{sec:future}.)

When we consider some agent $A$ w.r.t.\ which some task is given and for which we want to infer good actions, while information may come from (sensory data of) a collection $C$ of other agents, then we refer to $A$ as \defi{target agent} and the agents in $C$ as \defi{source agents}.

\paragraph{Experimentation platform ``Malmo''.}

\begin{figure*}[t]
\centering
\begin{subfigure}{.3\textwidth}
  \centering
  \includegraphics[width=.8\linewidth]{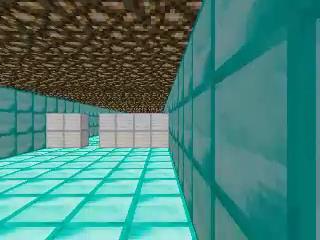}
  \caption{Mission 1}
  \label{fig::frame_exp1}
\end{subfigure}%
\begin{subfigure}{.3\textwidth}
  \centering
  \includegraphics[width=.8\linewidth]{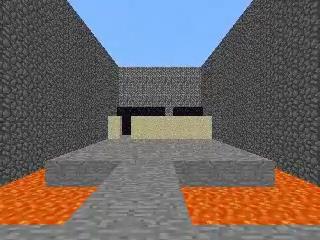}
  \caption{Mission 2}
  \label{fig::frame_exp2}
\end{subfigure}%
\begin{subfigure}{.3\textwidth}
  \centering
  \includegraphics[width=.8\linewidth]{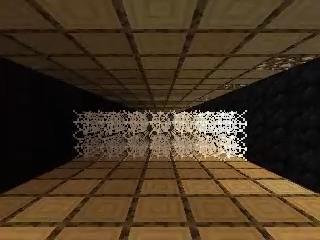}
  \caption{Mission 3}
  \label{fig::frame_exp3}
\end{subfigure}%
\caption{First frames of the respective missions, for illustration purposes.}
\label{fig:frame}
\end{figure*}

For the experiments in Section \ref{sec:aix_exp} we will use the software \defi{Malmo}, a simulated environment for experimentation with intelligent agents, that was introduced recently \cite{malmo2016}.
Malmo is based on ``Minecraft'', which is an open-ended computer game where players can explore, construct, collaborate, and invent their own ``games within the game'' or tasks. The Malmo platform provides an abstraction layer on top of the game through which one or more agents observe the current state of the world (observations are customizable) and interact with it through their specific action sets (or actuators). 
The advantage of Malmo is that it reflects important characteristics of the problem instances we will introduce in Section \ref{sec:problem}.
To illustrate the platform, three sample observations of an agent in three different maps in Malmo will be depicted in Figure \ref{fig:frame}.


\paragraph{Causal models.}
Mathematically, a \defi{causal model} \citep{Pearl2000,Spirtes2000} $M$ over a set $V$ of variables consists of a directed acyclic graph (DAG) $G$ with $V$ as node set, called \defi{causal diagram} or \defi{causal DAG}, and a conditional probability density $p_{X|\vn{PA}_X=pa_X}$ (for all $pa_X$ in the domain of $\vn{PA}_X$) for each $X \in V$, where $\vn{PA}_X$ are the parents of $X$ in $G$. 
Given a causal model $M$ and a tuple of variables $Z$ of $M$, the \emph{post-interventional causal model} $M_{\dc Z=z}$ is defined as follows:
 drop the variables in $Z$ and all incoming arrows from the causal diagram, and fix the value of variables in $Z$ to the corresponding entry of $z$ in all remaining conditional densities.
Based on this, we define the \defi{post-interventional density of $Y$ after setting $Z$ to $z$}, denoted by $p_{Y|\dc Z=z}$ or $p_{Y|\dc z}$, by the the density of $Y$ in $M_{\dc Z=z}$. 

On a non-mathematical level, we consider $M$ to be a correct causal model of some part of reality, if it correctly predicts the outcomes of \defi{interventions} in that part of reality (clearly there are other reasonable definitions of causation).
Keep in mind that in this paper, in particular Section \ref{sec:causal}, will use causal models and causal reasoning in a more intuitive and sometimes less rigorous way, to not be limited by the expressive power of the current formal modeling language.

Note that we will use expressions like $p(x|y)$ as shorthand for $p_{X|Y}(x|y)$.

%
%
%
%
%
%
%
%


\section[Problem formulation]{Problem formulation}
\label{sec:problem}


Let us describe the problem we consider in this paper in more detail:
\begin{itemize}
\item \emph{Given:} a task $T$ w.r.t.\ some partially unknown environment $E$, and additional heterogeneous but well-structured information sources $H$ (e.g., in the form of low-level sensory data, or in the form of high-level descriptions). 
\item \emph{Goal:} design an agent 
$A$ that automatically harnesses as much relevant information of $H$ as possible to solve $T$; more specifically, it should use $H$ to either improve an explicit model of the effects%
\footnote{In this sense, at least the target of the information integration is clear: modeling the dynamics or causal structure of the agent in the environment.}
of its actions, which then guides its actions, or let its actions directly be guided by $H$.%
\end{itemize}
Note that alternatively, one could also formulate the problem by letting $A$ only be an actuator, and not a complete agent, and include the agent's sensors into $H$. This might be a more precise formulation, however, for the sake of an intuitive terminology, we stick to the definition based on $A$ being an agent.%
\footnote{Note that the problem we formulate here does not coincide with developing (``strong'') artificial intelligence (AI), as defined, e.g., by the Turing test or simply based on human-level intelligence.
We restrict to sources of information that are more or less well-structured - either quantitative measurements with a simple and clear relation to the physical world, or information in a language much more restrictive than natural language.
Nonetheless, the formulated problem can be seen as one step from say RL into the direction of AI.
}

To illustrate the general problem, in what follows, we give three concrete examples of desirable scenarios in which agents automatically integrate heterogeneous information.
Ideally, agents would be able to \emph{simultaneously integrate information sources from all three examples}.



\subsection[Example 1]{Example 1: sharing information between different self-driving cars}
\label{sec:expl_vehicle}
\label{sec:expl_cars}

Consider self-driving cars.
It is desirable that as much information about the environment can be shared amongst them.
By such information we mean up-to-date detailed street maps, traffic information, information on how to avoid accidents etc.
For instance, assume that one self-driving car leaves the road at some difficult spot due to some inappropriate action, since, for instance, the spot has not been visited by self-driving cars before (or newly appeared due to say some oil spill or rockfall).
If we only consider other self-driving cars of the same hardware, this experience could directly be transferred to them by enforcing them not to perform the very action at the very spot.
(I.e., for all cars of the same hardware one could treat the experience as if it was their own and make them ``learn'' from it in the usual reinforcement learning (RL) way.)
However, if we assume that there are self-driving cars of different types, then it is not possible to transfer the experience, and thus avoid further accidents, in this straight-forward way.


\subsection[Example 2]{Example 2: observing another agent}
\label{sec:expl_obs_other}

Consider domestic robots.
A domestic robot may, with its sensor, observe humans how they handle doors, windows, light switches, or kitchen devices.
It should be possible that domestic robots learn from such experience.
For instance, one could imagine a robot to reason that, if it is able to operate the door knob in a similar way as a human did before, this would also open the door and and thus allow the robot to walk into the other room (to achieve some task).

%
%
%
%

\subsection[Example 3]{Example 3: integrating high-level information}
\label{sec:expl_new_city}

Consider an agent that arrives in a city it has never been to before.
The goal is to get to a certain destination, say to the town hall.
A resident may be able to explain the way in a simple language, with words such as ``... follow this street until you come to a church, then turn right ...''.
Or a resident could provide a map and mark directions on the map. 
One could imagine that an autonomous agent could combine such a description with a model of the ``local'' (or ``low-level'') dynamics that is shared by most environments (which is closely related to the laws of physics). 
The model of the ``local'' dynamics could have been either hard-coded, or inferred based on exploration in other (related) environments.
In principle it should be possible that such a combination allows the agent to successfully navigate to the destination city hall.

\section{Related work: potentials and limitations}
\label{sec:related}

Various research directions exist that address major or minor aspects of the problem formulated in Section \ref{sec:problem}.
Here we discuss the most relevant such directions we are aware of, highlighting their potentials and limitations w.r.t.\ the problem.
Keep in mind that Sections \ref{sec:analysis} and \ref{sec:causal_discussion} contain additional discussions on the advantages of our approaches over these directions.

\subsection{Reinforcement learning}
\label{sec:rl}

One of the most powerful approaches to shaping intelligent autonomous agents is \defi{reinforcement learning (RL)} \citep{sutton1998reinforcement}.
Instead of explicitly hard-coding each detail of an agent, for each environment and objective individually, the idea is to take an approach which is more modular and based on learning instead of hard-coding: the supervisor only determines the reward function and then the agent ideally uses exploration of the unknown environment and exploitation of the gained experience (sensory data) to achieve a high cumulative reward.

%
%

Regarding the problem we consider in this paper, RL plays a key role for integration of information in the form of \emph{recordings of an agent's own past}, or of an agent with the same hardware.
However, as mentioned in Section \ref{sec:problem}, in contrast to RL, here we consider the problem of integrating information beyond such recordings, such as sensory data from agents with different hardware, or higher level information such as maps.

\subsection{Learning from demonstrations}
\label{sec:lfd}

According to \citet{Argall2009}, in \defi{learning from demonstrations (LfD)}, some ``teacher'' performs a trajectory which is recorded, and the goal is that a ``learner'' agent, based on this recording, infers and imitates (or utilizes) the teacher's ``policy'' (or the dynamics of the environment, or both).
Central notions that \citet{Argall2009} uses to analyze and distinguish various types of LfD problems are the \defi{record mapping}, i.e., what aspects of the teacher's demonstration are measured and recorded, and the \defi{embodiment mapping}, i.e., if the recorded actions can directly be implemented by the ``learner'' and lead to similar observations as the recoded ones, or if the recordings first have to be transformed ``to make sense'' for the learner. 

Our problem formulation can be seen as a generalization of LfD. 
Based on this, while a significant part of the problem we consider can be addressed by LfD methods, others are beyond the scope of these methods:
Instead of hand-crafting, e.g., the embodiment mapping for each agent individually, we aim at (semi-)automating the inference of the mapping from recordings of ``source'' agents to actions of a ``target'' agent.
In particular, we propose to do such a (semi-)automation based on additional information sources on the hardware specifications of the agents involved (Section \ref{sec:causal}).%
\footnote{Note that there is some work on learning from observations only (not actions) of a ``teacher'' \citep{Price2003}. However, this approach does not allow to integrate information such as the map in \nameref{sec:expl_new_city}. Note that a difference to our method in Section \ref{sec:aix_exp} is that, e.g., an estimate of the complete transition probability is necessary, while our method only requires an idea of the ``low-level'' dynamics.} 

Generally, we aim at integrating information from many different sources simultaneously (e.g., many other self-driving cars in \nameref{sec:expl_cars} and many different forms of information as described in \nameref{sec:expl_cars} through \nameref{sec:expl_new_city}.
In particular, we aim at learning from databases that contain desirable as well as undesirable trajectories (e.g., avoid similar accidents as in \nameref{sec:expl_cars}).

Clearly, we do not present methodology that fully tackles the above shortcomings of LfD methods in this paper. Rather, we make first steps towards such methodology in Sections \ref{sec:aix_exp} and \ref{sec:causal}.



%

\subsection{Multi-agent systems}
\label{sec:mas}

In multi-agent systems, collections of agents acting in a shared environment are studied \citep{stone2000multiagent}.
One important task is collaboration between agents \citep{olfati2007consensus}.
A common approach is to model the collection of agents again as a single agent, by considering tuples of actions and observations as single actions and observations.
Learning-based methods have been extensively studied \citep{bucsoniu2010multi}.

While multi-agent systems approaches often allow to share and transfer information between agents, regarding the problem we formulated in Section \ref{sec:problem} they have certain limitations:
Similar as LfD, they usually do not integrate higher-level information sources (as the map in \nameref{sec:expl_new_city}) or explicit hardware specifications of the agents (which we do in Section \ref{sec:causal}).
Furthermore, if the mapping from some source agents sensory data to a target agents action is learned via modeling all agents as a single one, then it seems difficult to add agents to an environment, while our preliminary investigation in Section \ref{sec:causal} in principle allows for adding agents more easily.
Also note that the task of collaboration between agents is rather external to the problem we consider.

%
%
%
%
%
%
%

%
%
%

\subsection{Transfer learning for agents}
\label{sec:tl}

The problem we consider is related to transfer learning for agents.
For instance, \citet{taylor2009transfer} consider an example where, in the well-known mountain car example, experience should be transferred although the motor of the car is changed.
This comes close to transferring experience between self-driving cars as we suggest in \nameref{sec:expl_cars}.
However, the scope of methods reviewed in \citep{taylor2009transfer} is on transferring observation-action recordings or things such as policies, value functions etc. using an appropriate mapping, while the goal we pursue is to also integrate information which is usually not expressible in these terms (e.g., the map or natural language description in \nameref{sec:expl_new_city}, or the observation of another agent in \nameref{sec:expl_obs_other}).
Furthermore, in this paper we aim at integrating \emph{many heterogeneous sources} of  information, while in transfer learning, even though several sources of information may be considered, they are usually homogeneous.

\subsection{Further related areas}

Other related directions include the following.
Recently, the experimentation with intelligent agents in platforms based on computer games has become popular \citep{mnih2015human}.
To our knowledge, the current work is the first one to use such platforms to study the problem of information integration, or related problems such as LfD (Section \ref{sec:lfd}).

The general integration and transfer of data (not focused on intelligent agents) using causal models has been studied by \citet{Pearl2011,bareinboim2015causal}.
The idea of integrating higher-level information (again not for intelligent agents though) has been studied, e.g., by \citet{vapnik2009new}.
The relation between intelligent agents and causal models has been studied from a more philosophical perspective, e.g., by \citet{woodward2005making}.

Another related areas is computerized knowledge representation \cite{sowa1999knowledge}.
Compared to general approaches to knowledge representation, our focus is on knowledge about the \emph{physical} world.

Another relevant area is integration of human knowledge \cite{holzinger2016interactive}.

\section{Investigation 1: integration of ``non-subjective'' information, evaluated in a simulated environment}
\label{sec:aix_exp}

In this section we aim to shed light on the following aspects of the problem formulated in Section \ref{sec:problem}:
\begin{itemize}
\item Generally, what experiments, in particular in simulated environments, can be performed to better understand the problem?
\item How can experimentation (exploration) help an agent to translate ``non-subjective'' experience not recorded by itself into its own ``coordinate system'' and use it for (successful) decision making?
\item How can partial information on the dynamics, such as a controller that is known to work locally, be merged with ``higher-level'' information such as hints on the path to some goal position?
\item How can we quantify the efficiency gain from additional information sources?
\end{itemize}

The investigation is structured as follows: 
in Section \ref{sec:scenario} we describe the scenario, in particular the available heterogeneous information sources,
in Section \ref{sec:method} we sketch an information-integrating agent for that scenario,
then, in Section \ref{sec:eval}, we evaluate an adapted version of the agent in a simulated environment, 
and last, in Section \ref{sec:analysis} we discuss the advantages of our method over those that do not use additional information, and some further aspects.



\subsection{Scenario}
\label{sec:scenario}

\paragraph{Task.}

An agent $A$ starts in some unknown landscape and the task is to get to some visually recognizable goal position as quickly as possible.

\paragraph{Available heterogeneous information.}

We assume the following information sources to be available:
\begin{itemize}
\item the agent's own sensory input in the form of images $y_t$ and position signal $q_t$ (which can be seen an ``interactive information'' source since the agent can ``query'' this source via its actions), 
\item the controller $\vn{ctl}$, which can be seen as a summary of the agent $A$'s past subjective experience regarding the invariant local ``physical laws'' of a class of environments%
\footnote{Specifically, we assume that if the distance between position $q_1$ and $q_2$ is small, then $\vn{ctl}$ successfully steers from $q_1$ to $q_2$.
Alternatively, $\vn{ctl}$ could be a local model of the dynamics which induces such a controller.},
\item a video trajectory $y^*_{0:L}$ that is a first-person recording of another agent with similar (but not necessarily identical) hardware that runs to the goal in the same environment. 
\end{itemize}

\paragraph{Relation to the problem formulated in Section \ref{sec:problem}.}
On the one hand, this scenario can be seen as a (very) simplistic version of the scenario described in \nameref{sec:expl_cars}:
$A$ is a self-driving car that is supposed to get to some marked goal in an unknown environment, and the video $y^*_{0:L}$ comes from other cars that have a similar video-recording device but  different hardware (engine etc.).

On the other hand, this scenario can be seen as a simplistic version of the scenario described in \nameref{sec:expl_new_city}:
the unknown landscape is some unknown city $A$ arrived in, and instead of a description of the way to the destination in simple natural language, it gets a sequence of photos that describe the path it has to take.

\subsection{Method}
\label{sec:method}

\begin{algorithm}[tb]
   \caption{Agent that integrates first-person video of other agent}
   \label{alg:simple_integration}
\begin{algorithmic}[1]
   \STATE \stress{input:} Controller $\vn{ctl}$, video $y^*_{1:L}$.
   \FOR{i = 1, \ldots, L} \label{code:for}
   \STATE Use local controller $\vn{ctl}$, optimization method $\vn{opt}$ and interaction with the environment to search locally around the current position for the next $q_i = \arg\min_q  \vn{dist}(y^*_i, \E(Y | Q=q))$.
   \STATE Use $\vn{ctl}$ to go to $q_i$.
   \ENDFOR \label{code:endfor}
\end{algorithmic}
\end{algorithm}


First we sketch a general method, i.e., ``software'' for $A$, in Algorithm \ref{alg:simple_integration}, assuming a (stochastic) optimization method $\vn{opt}$ and an image distance $\vn{dist}$ as given (for concrete examples, see below).
Note that in Algorithm \ref{alg:simple_integration} we denote by $\E(Y|Q=q)$ the mean image $Y$ observed at position $Q=q$.
The basic idea is that the agent uses local experimentation, based on prior knowledge of the local dynamics, to map the video $y^*_{1:L}$ into information (and eventually actions) that directly describes its own situation.

Although Algorithm \ref{alg:simple_integration} is in principle applicable to the experimental setup we consider in Section \ref{sec:exp} below, we will evaluate Algorithm \ref{alg:integration_implementation} instead, which is a simplified proof-of-concept implementation of it, making use of ``teleportation'', allowing the agent to directly jump to other positions without needing to navigate there.
For Algorithm \ref{alg:integration_implementation}, as optimization method $\vn{opt}$, we use simple grid search.
(Note that instead one could use gradient descent or Bayesian optimization techniques.)
Furthermore, we define the image distance $\vn{dist}$ using Gaussian blur as follows:
$
\vn{dist}(u, v) := \| N \ast (\bar{u} - \bar{v}) \|
$, where $\bar{u}$ is the normalization of $u$ (i.e., subtraction by mean and division by standard deviation over the single pixels), $N$ is a bivariate Gaussian with hand-tuned variance and $\ast$ is the convolution in both image dimensions.

\begin{algorithm}[tb]
   \caption{Proof-of-concept of Algorithm \ref{alg:simple_integration} for Malmo}
   \label{alg:integration_implementation}
\begin{algorithmic}[1]
   \STATE \stress{input:} Controller $\vn{ctl}$, video $y^*_{1:L}$.
   \STATE set $r_0 = $ current position, once the mission starts
   \FOR{i = 1, \ldots, L}  \label{code:for}
   \STATE use $\vn{ctl}$, $\vn{opt}$ and teleportation to locally search around position $r_{i-1}$ for the next $r_i = \arg\min_r \E( \vn{dist}(y^*_i, Y) | Q=r)$
   \ENDFOR
   \STATE restart the mission
   \STATE set $i:=0$.
   \WHILE{$i < L$} \label{code:while}
   \STATE use $\vn{ctl}$ to steer to  $r_{i}$
   \IF{current position is close to $r_i$}
   \STATE{set $i := i+1$}
   \ENDIF
   \ENDWHILE
\end{algorithmic}
\end{algorithm}

\subsection{Empirical evaluation in a simulated environment}
\label{sec:exp}
\label{sec:eval}
\label{sec:setup}
\label{sec:outcome}

\begin{figure}[t]
\centering
\begin{subfigure}{.33\columnwidth}
  \centering
  \includegraphics[width=0.8\linewidth]{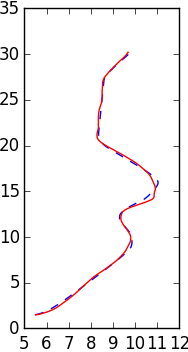}
  \caption{Mission 1}
  \label{fig::traj_exp1}
\end{subfigure}%
\begin{subfigure}{.33\columnwidth}
  \centering
  \includegraphics[width=0.8\linewidth]{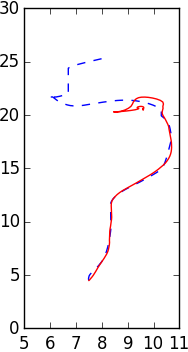}
  \caption{Mission 2}
  \label{fig::traj_exp2}
\end{subfigure}%
\begin{subfigure}{.33\columnwidth}
  \centering
  \includegraphics[width=0.8\linewidth]{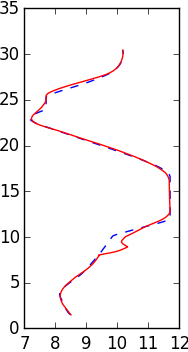}
  \caption{Mission 3}
  \label{fig::traj_exp3}
\end{subfigure}%
\caption{The ``ground truth'' position trajectory of the demonstrator $q^*_{0:L}$ (blue dashed line), and the position trajectory of Algorithm \ref{alg:integration_implementation} $\hat{q}_{0:K}$, (red solid line) from top view (x- and y-axis correspond to x- and y-coordinate in the map. While Algorithm \ref{alg:integration_implementation} fails in Mission 2 due to the repetitive structure of some wall, it succeeds in Missions 2 and 3 in spite of its simplicity.}
\label{fig::traj}
\end{figure}

\begin{table}
\caption{For each mission (row) a short description (column 2), and the outcome of Algorithm \ref{alg:integration_implementation} applied to it (columns three and four).}
\label{table::exp}
\begin{center}
\begin{tabularx}{\columnwidth}{l|X|ll}
\bf Mission & \bf Description and image & \bf $U$ & \bf $\hat{q}_{0:K}$ versus $q^*_{0:L}$ \\
\hline \\
1 & Figure \ref{fig::frame_exp1}. Two passages have to be passed. & success & Figure \ref{fig::traj_exp1} \\
2 & Figure \ref{fig::frame_exp2}. With (mortal) lava next to the path. & fail & Figure \ref{fig::traj_exp2} \\
3 & Figure \ref{fig::frame_exp3}. With spider webs that occlude vision and lead to slow motion. & success & Figure \ref{fig::traj_exp3}
\end{tabularx}
\end{center}
\end{table}


To evaluate Algorithm \ref{alg:integration_implementation}, we consider three simple ``Parkours'' missions in the experimentation platform Malmo, described in Section \ref{sec:pre}. 
These missions consist of simple maps that have a special, visually recognizable, position which is defined as goal.
A short description of the three missions is given in column two of Table \ref{table::exp}.
We generally restrict the possible actions to $[-1, 1]^2$, where the first dimension is moving forth and back, and the second is strafing (moving sideways).
The \emph{task} is to get to the goal position within 15 seconds in these maps. 

For each mission, we record one trajectory performed by a human demonstrator, which solves the task. 
More specifically, we record positions, which we denote by $q^*_{0:L}$, and observations (video frames), denoted by $y^*_{0:L}$. 

We run Algorithm \ref{alg:integration_implementation} with inputs $y^*_{0:L}$ and a simple proportional controller \citep{astrom2010feedback} (for $\vn{ctl}$), where we tuned the proportional constant manually in previous experiments (but without providing $q^*_{0:L}$ or the actions the human demonstrator took).
Let $\hat{q}_{0:K}$ denote the trajectory of positions that Algorithm \ref{alg:integration_implementation} subsequently takes in the map. 
Furthermore, let $U \in \{\text{fail}, \text{success}\}$ denote whether the position tracking while-loop of Algorithm \ref{alg:integration_implementation} (line \ref{code:while} and following) gets to the goal within 15 seconds.
(This is a significantly weaker evaluation metric than considering the runtime of the complete Algorithm \ref{alg:integration_implementation} of course.)


The outcome of the experiment is given in columns three and four of Table \ref{table::exp} and Figure \ref{fig::traj}.

%
%
%



\subsection{Discussion} 
\label{sec:exp_disc}
\label{sec:analysis}



%
%
%
%
%

\subsubsection{Discussion of the experiment}

\paragraph{Outcome.}
As shown by Table \ref{table::exp} and Figure \ref{fig::traj}, Algorithm \ref{alg:integration_implementation} is successful for Missions 1 and 3. It fails in Mission 2 due to the repetitive structure of a wall that fills the complete image that is observed at some point during $y^*_{1:L}$.
This wall makes the mapping from position to observation (video frame) (locally) non-injective which makes the algorithm fail. 
Note that this problem could quite easily be overcome by using prior assumptions on the smoothness of $q^*_{0:L}$ together with considering more than one minimum of $\vn{dist}$ as the potential true position (using, e.g., Bayesian optimization) or by searching for position sequences longer than 1 that match $y^*_{1:L}$.

\paragraph{Limitation of the experiment.}
A clear limitation of the experiment is that the human demonstrator that produced $y^*_{1:L}$ used the same (simulated) ``hardware'' as Algorithm \ref{alg:integration_implementation}, while the overall goal of this paper is to integrate heterogeneous information.
However, we hope that this experiment can form the basis for more sophisticated ones in the future.

\subsubsection{Theoretical analysis and insight}

\paragraph{Efficiency gain from harnessing $y^*_{1:L}$.}

\label{sec:efficiency_gain}
Assuming there are at most $N$ positions which the demonstrator can reach within one time step, Algorithm \ref{alg:simple_integration} takes only about $O(L \cdot N)$ steps to get to the goal.
Note that this theoretical analysis is supported by the empirical evaluation: Algorithm \ref{alg:integration_implementation}'s trajectories - visualized in Figure \ref{fig::traj} - are roughly as long as $y^*_{1:L}$ (note that the visualization does not show the local search of length $N$).

This has to be contrasted with an agent that does not integrate the information $y^*_{1:L}$, and therefore, in the worst case, has to search \emph{all} positions in the map, a number which is usually is much higher than $O(L \cdot N)$ (roughly $O(L^2)$).

\paragraph{Comparison to LfD.}
The task we study is closely related to LfD.
However, note that usually in LfD, the target agent (``learner'') has access to the demonstrator's actions, which is not necessary for our method.
Furthermore, in our method, in some sense, the target agent can be seen as translating $y^*_{1:L}$ into its own ``coordinate system'' itself, while this mapping is usually hand-crafted in LfD.

\paragraph{Some insights during the development of the method.}
An interesting insight during the development of Algorithm \ref{sec:method} was that while the low-pass filter only led to minor improvements, what really helped was to visit each position say three times and then optimize the distance over the \emph{averaged} images.
Furthermore, it was was surprising how well the simple Euclidean distance we used worked.

\paragraph{Limitations.}

Note that the method is limited to environments similar to landscapes where some stochasticity and variation may be contained, but not too much.
For instance, if the environment varies to strongly in the dependence on the time an agent spends in the environment, the proposed method most likely fails since the tracking of $y^*_{1:L}$ usually takes longer than the original performance of it.

\section{Investigation 2: integrating sensory data, hardware specifications and causal relations}
\label{sec:causal}

In this section we aim to shed light on the following aspects of the problem formulated in Section \ref{sec:problem}:
\begin{itemize}
\item How can information on the the hardware specifications of various agents be used for knowledge transfer between them?
\item To what extent can \emph{causal models} help, e.g., for integrating those hardware specifications (i.e., information on the ``data producing mechanisms'')?\footnote{Another reason why causal reasoning could help is that in the end we are interested in the \emph{causal effects} of an agent on its environment, and not just correlational information.} 
\item How can information from the \emph{``subjective perspective''} of an agent (i.e., on the relation between its sensory measurements and its actions) be merged with information from an ``outside perspective'' (i.e., that of an engineer which sees the hardware specifications of an agent).
\end{itemize}

The investigation is structured as follows: 
in Section \ref{sec:causal_scenario} we describe the scenario, in particular the available heterogeneous information sources,
in Section \ref{sec:causal_method} we outline an information-integrating agent for that scenario,
then, in Section \ref{sec:causal_example}, we give an intuitive toy example of scenario and method,
and last, in Section \ref{sec:causal_discussion} we discuss advantages and limitations of our approach.

Keep in mind the definition of causal models in Section \ref{sec:pre}.
It needs to be mentioned that at certain points in this section we will allow ourselves some extent of imprecision (in particular in the treatment of the (causal) model $M$ and how it is inferred), since  we aim at going beyond what current rigorous modeling languages allow.


\subsection{Scenario}
\label{sec:causal_scenario}

\paragraph{Task.}

We consider a scenario where a collection $C$ of autonomous agents, think of self-driving cars, operates in a shared environment.
(For simplicity we assume that the number of agents is small compared to the size of the environment, such that they do not affect each other.)
We assume that while some hardware components of the agents differ, others are \emph{invariant} between them.
We assume that for each car a task (e.g., to track some trajectory) is given and fixed.

\paragraph{Available heterogeneous information.}

Note that we could allow $C$ to vary over time, e.g., to account for the fact that new cars get on the road every day, however, for the sake of a simple exposition, we leave it fixed here.
We assume the following information sources to be available at time $t$:
\begin{itemize}
\item specifications $\vn{spec}^k$ of the hardware of each agent $k$ 
\item past experiences (i.e., actions and observations) $e^k_{t}$ of all agents $k \in C$, consisting of observations $y^k(t)$ and control outputs $u^k(t)$, i.e., $e^k_{t} = (u^k(0), y^k(0), \ldots, u^k(t), y^k(t))$.
\item a description $D$ (e.g., a physical or causal model or collection of such models) consisting of (1) a set of \emph{independence statements}%
\footnote{Clearly, independence assertions are central to integration of information: only based on statements of the form ``$Y$ is more or less independent of all factors that potentially will be included, except for this and this small set'' it seems possible to rigorously (automatically) reason about integration.}
and (2) a set of dependence statements, possibly including specific information on the shape of the dependence, w.r.t.\ the factors contained in the specifications $\vn{spec}^k$ and the experiences $e^k$, for $k \in C$. 
Potentially, the various independence and dependence information pieces could come from different sources, say targeted experiments as well as general prior knowledge.
We assume the dependence structure (including the precise shapes of the dependences) to be time-invariant.
\end{itemize}

\paragraph{Relation to general problem.}
This scenario captures certain aspects of \nameref{sec:expl_new_city} in that some higher-level information in the form of the description $D$ is available.
While here, as a first step, we only consider mathematical models, in the future one could also imagine to include informal but well-structured models and descriptions (in simple natural language), possibly translating them into formal models as an intermediate step (using, e.g., machine learning).
Furthermore, the scenario captures important aspects of \nameref{sec:expl_cars}, since we consider the integration of information from certain source self-driving cars for the decision making of a given target car.

\subsection{Sketch of a method}
\label{sec:causal_method}

\begin{algorithm}[tb]
   \caption{Integration and control algorithm for agent $j$}
   \label{alg:higher_level}
\begin{algorithmic}[1]
   \STATE \stress{input:} Time point $t$, description $D$, specifications $(\vn{spec}^k)_{k \in C}$, experiences $(e^k_{t})_{k \in C}$.
   \STATE Initialize a causal model $M$ by the causal diagram implied by the description $D$ over the set of factors in $(\vn{spec}^k)_{k \in C}$ and $(e^k_{t})_{k \in C}$. \label{code:M_from_D} 
   \STATE Update the ``belief'' over the mechanisms in $M$ using all values of variables contained in $(\vn{spec}^k)_{k \in C}$ and the experiences $(e^k_{t})_{k \in C}$ (possibly based on additional priors). \label{code:update_M} 
   \STATE From the updated $M$, calculate $M^j$, the implication of $M$ for agent $j$. \label{code:update_j}
   \STATE Find action $u(t)$ that is optimal w.r.t.\ the given task, under $M^j$. \label{code:act}
   \STATE \stress{output:} $u(t)$
\end{algorithmic}
\end{algorithm}

We sketch a method for the described scenario in Algorithm \ref{alg:higher_level}.
It first derives a ``global'' causal diagram - applying to all agents - from the potentially heterogeneous description $D$ (line \ref{code:M_from_D}).
Then (line \ref{code:update_M}) the causal conditionals of $M$, which are not determined by $D$, are inferred from the given hardware specifications as well as the experiences gathered by all agents up to time $t$.
Last, based on the hardware specifications of agent $j$, the implications of $M$ for agent $j$ are calculated (line \ref{code:update_j}) and the optimal action under these implications is performed (line \ref{code:act}).

%
%
%

%
%
%
%
%

\subsection{A toy example}
\label{sec:causal_example}

Let us illustrate how the method proposed in Section \ref{sec:causal_method} works in a concrete toy scenario.
The core intuition is that while some details of the dynamics of self-driving cars may vary between different cars, they can still share information on say the road conditions (friction, drag, etc.) at certain positions $y$, or the like.

\begin{figure}
\centering 
\begin{tikzpicture}[scale=0.6]
    
\node[obs] at (0, 0) (ddx) {$\ddot{y}(t)$};

\node[obs] at (-1, 2) (A_e) {$F(t)$} edge[->] (ddx);
\node[obs] at (1, 2) (A_f) {$G(t)$} edge[->] (ddx);

\node[obs] at (-1, 4) (u) {$u(t)$} edge[->] (A_e);
\node[obs] at (1, 4) (x) {$y(t)$} edge[->] (A_f);

\node[obs] at (-3, 2) (m) {$hp$} edge[->] (A_e);


\end{tikzpicture}
\caption{Sketch of the causal diagram $H$. The power $\vn{hp}$ influencing only $F(t)$ implies that knowledge on the mechanism $f_G$ for $G(t)$ can be transferred between two cars even if they differ in $\vn{hp}$.}
\label{fig:diagram}
\end{figure}
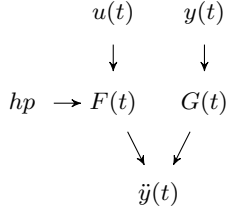

\paragraph{Specific scenario.}

We consider two simplified self-driving cars, i.e., $C = \{1, 2\}$, where we assume the observation $y^k(t)$ (or $y(t)$ if we refer to a model for both cars) to be the car's position.
We consider the following concrete instances of the information sources listed in Section \ref{sec:causal_scenario}.
(Note that, for the sake of simplicity, we assume all mechanisms to be deterministic, such that we can model them by functions $f_{\ldots}$ instead of conditional distributions $p(\ldots|\pa_{\ldots})$, although the latter would be more general, see Section \ref{sec:pre}.)
\begin{itemize}
\item The specification for car $k$ is given by its power (e.g., measured in horse powers), i.e., $\vn{spec}^k = \vn{hp}^k$. 
\item The experience of car $k$ consist of all past position-action pairs of both cars, i.e., $e^k_t = (u^k(0), y^k(0), \ldots, u^k(t), y^k(t))$.
\item The description $D$ consists of three elements:
\begin{itemize}
\item an engineer contributes the equation $F(t) = f_{F}(u(t), \vn{hp})$ for the engine of the cars\footnote{In particular, the mechanism specification implies non-influence by all other relevant factors.}, where $F(t)$ is the force produced by the engine (incl.\ gears), and $f_{F}$ a known function,
\item a physicist contributes the equation $\ddot{y}(t) = \frac{1}{m}( F(t) + G(t) )$ for the acceleration $\ddot{y}(t)$ of the cars, where $G(t)$ are other forces that affect the cars, such as friction and drag, and $m$ is the known mass (which we assume to be the same for both cars),
\item another physicist produces a set of additional independence assertions, such that altogether the description $D$ implies the causal diagram $H$ depicted in Figure \ref{fig:diagram}.\footnote{A better way to describe the physical forces causally might be to replace $\ddot{y}(t)$ in the equations and in $D$ by expressions based on $v(t + \Delta t)$, $v(t)$, and $\Delta t$, where $v(t)$ denotes the velocity.} 
\end{itemize}
\end{itemize}


\paragraph{Implementation of our method.}

We suggest the following concrete implementation of the crucial part of our method, i.e., line \ref{code:update_M} and \ref{code:update_j} in Algorithm \ref{alg:higher_level}, based on the concrete instances of information available in this toy scenario.
Keep in mind that $f_G$, the function that maps a position $y$ to the corresponding force $G$ at that position and thus models the generating mechanism for $G(t)$, is the only unknown part of $M$ after initializing it by $D$. 
\begin{itemize}

\item Line \ref{code:update_M}: Use the experience $(e^k_{t})_{k \in C}$, to infer the function $f_{G}$ on all positions $y$ visited by either of the cars, based on the equation
\begin{align*}
m \ddot{y} - f_{F}(u,\vn{hp}^k) = f_{G}(y)
\end{align*}
for all $k \in C$, and the fact that the l.h.s.\ of this equation as well as $y$ are known for all positions (and accelerations) visited by either of the cars.
\item Line \ref{code:update_j}: calculate ``$p(\ddot{y} | do(u), y, \vn{hp^j})$'', i.e., the effect of control action $u$ of agent $j$ at position $y$, \emph{for all positions that were visited before by either of the cars}.
\end{itemize}

\subsection{Discussion}
\label{sec:causal_discussion}


The toy example in Section \ref{sec:causal_example} shows how in principle the integration of heterogeneous information could help some ``target'' self-driving car for better decision making in situations not visited by it but by different ``source'' self-driving cars.
It is important to note that all listed information sources were necessary for this:
the hardware specifications are necessary to understand $F(t)$,
the experience is necessary to infer $f_G$,
and the independence knowledge ($hp$ not affecting $G(t)$) is necessary to transfer the knowledge about the force $G(t)$ on various positions $y$ between the cars.
Note that the above scenario cannot be tackled by standard RL approaches since we transfer knowledge between agents of different hardware. 
Furthermore, methods like LfD or transfer learning (see Sections \ref{sec:lfd} and \ref{sec:tl}) usually do not automatically harness information on hardware specifications of agents.

Based on our preliminary investigation above, it seems that causal models are helpful in that they provide a language in which one can express relevant assumptions and reason about them.
However, from a practical perspective, it is not clear if the necessary calculations could not be genuinely done e.g. in classical probabilistic models.

An important question is whether the method sketched in Section \ref{sec:causal_method} can be generalized to dependence statements in more natural - but still well-structured - language than equations and causal diagrams.

\section{Outlook: future directions and open questions}
\label{sec:future}

Here we sketch a potential agenda for future investigations and pose interesting open questions.

\subsection{Potential future directions}
\label{sec:future_ideas}

\paragraph{``Universal representation of physical world''.}

An interesting subject-matter of future research would be a ``universal representation'' of the physical world 
- a rich representation to which each information source could be translated, and from which each agent could derive the implications for its specific sensor and actuator configurations.
Such a representation would be more efficient than hand-crafting mappings for each (new) pair of source and agent individually (as is usually done in e.g.\ LfD, see Section \ref{sec:lfd}),
reducing the number of necessary mappings from $n^2$ to $n$, where $n$ is the number of agents. 
One starting point would be representations that are already used to integrate laser or radar scanner data on the one hand with (stereo) video camera data on the other hand in self-driving cars \cite{geiger2011stereoscan}.
Another starting point would be the global positioning system (GPS) which is a successful universal representation of position with clear, hardware-independent semantics.

\paragraph{Investigation and classification of the ``integration mapping''.}

Another important concept for integration of heterogeneous information could be the \emph{mapping} that transforms a collection of pieces of well-structured heterogeneous information into a model of the current situation or even directly into action recommendations.
The study of such a mapping could build on the investigation of related mappings in LfD and transfer learning for agents (see Sections \ref{sec:lfd} and \ref{sec:tl}).
Furthermore, parts of such a mapping could be learned (which would be related to machine-learning-based multi-agent systems, see Section \ref{sec:mas}.
Generally, it would be interesting to examine the basic conditions under which the integration of heterogeneous information can be beneficial.
Note that one way to \emph{classify and order} various sources of information (and the mappings that are necessary to integrate them) would be from ``closest to the agent $A$'', i.e., it's own past observations and actions to ``most distant'', e.g., agent-independent descriptions of the world in simple natural language, as exemplified in Section \ref{sec:problem}. 

\paragraph{Further experiments.}
Further experiments, with a gradually increasing difficulty (e.g.\ along the ordering proposed in the previous paragraph), could be performed, e.g. using the platform Malmo (Section \ref{sec:pre}) to gain a better understanding of integration of information:
\begin{enumerate}
\item Agents can observe other agents from a third-person perspective, enabling \nameref{sec:expl_obs_other} in Section \ref{sec:problem}.
\item Higher level observations can be provided in the form of natural language (typed chat or external information in the form of natural language), or through artifacts such as maps, sign-posts, symbolic clues, etc. 
\end{enumerate}

\subsection{Open questions}

It would also be interesting to investigate how the following questions could be answered:
\begin{itemize}
\item One of the main question which guided our investigation in Section \ref{sec:causal} can be cast as follows. While the information relevant to an agent is usually in the form of effects of its actions in certain situations, a lot of knowledge is formulated in non-causal form: for instance street maps at various granularities for self-driving cars. How are these two forms of information related? Is there a standard way to translate between them?
Stated differently, how can various forms of information be translated into a model of the \emph{dynamics} of the agent in the world.
\item Where is the boundary between additional heterogeneous information and prior knowledge?
\item How can the need for information integration be balanced with \emph{privacy} restrictions? For instance, one may imagine cases where the mapping from a source agent's experience to a target agent's action is rather simple in principle, but information collected by the source agent cannot or should not be transmitted to the other, at least not in full.
\item How can big databases of information be filtered for useful information, i.e., the information which is correct and relevant for the current environment and task?
\item To what extent is the problem of information transfer between two different agents in the ``same'' environment just a special case of transfer between different environments (by considering the hardware of an agent as part of the environment)?
\item How can we reason without having a \emph{``global'' model} such as Figure \ref{fig:diagram}? What about interfaces to build global from \emph{``local'' models}, describing only say the engine?
\item Generally, what are potential theoretical limitations of automated information integration, e.g.\ in terms of computability?

\end{itemize}

\section{Conclusions}
\label{sec:conclusions}

In this paper, we considered the problem of designing agents that autonomously integrate available heterogeneous information about their environment.
We investigated how experimentation in simulated environments on the one hand, and causal models on the other, can help to address it.
A next step would be to perform more sophisticated experiments, ideally guided by specific problems e.g.\ from the area of self-driving cars.

\section{Acknowledgments}

The authors thank Mathew Monfort, Nicole Beckage, Roberto Calandra, Matthew Johnson, Tim Hutton, David Bignell, Daniel Tarlow, Chris Bishop and Andrew Blake for helpful discussions, and the anonymous reviewers for useful hints.

\bibliography{master_include/master_bibfile,additional_include/integration}

\end{document}

%% file: master_include/master_preamble.tex
\usepackage{pgfplots}
\usepackage{paralist}

\usepackage[utf8]{inputenc}
\usepackage{amsmath,amssymb,amsthm}
\usepackage{tikz}
\usetikzlibrary{arrows}
\usepackage{hyperref}

\usepackage{algorithm}
\usepackage{algorithmic}

\newcommand{\stress}{\textbf}

\newcommand{\defi}{\emph}

\newcommand{\E}{\mathbb{E}}

\newcommand{\dc}{\mathrm{do}\,}

\newcommand{\pa}{\mathrm{pa}}

\newcommand{\vn}[1]{\mathit{#1}} 

\definecolor{light-gray}{gray}{0.75}

\usetikzlibrary{arrows,shapes,plotmarks}
\tikzset{>=stealth'} 
\tikzstyle{graphnode} = [circle,draw=black,minimum size=22pt,text centered,text width=22pt,inner sep=0pt] 
\tikzstyle{var}   =[graphnode,fill=white]
\tikzstyle{const}   =[graphnode,fill=white,draw=none]
\tikzstyle{hid}   =[graphnode,fill=gray,draw=none,text=white]
\tikzstyle{phantom}   =[graphnode,fill=white,draw=none,text=white]
\tikzstyle{halfhid}   =[graphnode,fill=light-gray,draw=black,text=white]
\tikzstyle{obs}   =[graphnode,fill=none,draw=none]
\tikzstyle{sel}   =[rectangle,fill=none,draw=black]
\tikzstyle{selection}   =[rectangle,draw=none,fill=black,minimum size=5pt,inner sep=0pt,outer sep=0pt]
\tikzstyle{fac}   =[rectangle,draw=black,fill=black!25,minimum size=5pt]
\tikzstyle{facprior} =[rectangle,draw=black,fill=black,text=white,minimum size=5pt]
\tikzstyle{edge}  =[draw=white,double=black,thick,-]
\tikzstyle{prior} =[rectangle, draw=black, fill=black, minimum size=5pt, inner sep=0pt]
\tikzstyle{dirprior} = [circle, draw=black, fill=black, minimum size=5pt, inner sep=0pt]